\documentclass[letterpaper, 10 pt, conference]{ieeeconf}

\IEEEoverridecommandlockouts

\usepackage{graphics}
\usepackage{epsfig}
\usepackage{mathptmx}
\usepackage{times}
\usepackage{amsmath}
\usepackage{amssymb}
\usepackage{amsmath,amsfonts}
\usepackage{algorithmic}
\usepackage{algorithm}
\usepackage{array}
\usepackage{cleveref}
\usepackage{textcomp}
\usepackage{stfloats}
\usepackage{url}
\usepackage{verbatim}
\usepackage{graphicx}
\usepackage{cite}
\usepackage{tikz}
\usepackage{subcaption}
\usepackage{graphicx}
\usepackage{makecell}
\usepackage{tabularx}
\usepackage[table]{xcolor}

\title{
\textbf{ANGEL}: \textbf{A} \textbf{N}ovel \textbf{G}ripper for V{\textbf{E}rsatile} and \textbf{L}ight-touch Fruit Harvesting}

\author{Dharmik Patel$^{1}$, Antonio Rafael Vazquez Pantoja$^{1}$, Jiuzhou Lei$^{1}$, \\ Kiju Lee$^{2}$, Xiao Liang$^{3}$, Minghui Zheng$^{1}$
\thanks{This work was supported by the Texas A\&M AgriLife Institute for Advanced Health Through Agriculture Transformative Research Capacity Funding.}
\thanks{$^{1}$ D. Patel, A. Pantoja, J. Lei, and M. Zheng are with J. Mike Walker ’66 Department of Mechanical Engineering, Texas A\&M University, 
{\tt\footnotesize\{dharmik.p45, av2006, jiuzl, mhzheng\}@tamu.edu}. }
\thanks{$^{2}$ K. Lee is with Department of Engineering Technology \& Industrial Distribution and J. Mike Walker ’66 Department of Mechanical Engineering, Texas A\&M University, 
{\tt\footnotesize\{kiju.lee\}@tamu.edu}. }
\thanks{$^{3}$ X. Liang is with Zachry Department of Civil and Environmental Engineering, Texas A\&M University, 
{\tt\footnotesize\{xliang\}@tamu.edu}. }
}

\begin{document}

\maketitle
\thispagestyle{empty}
\pagestyle{empty}

\begin{abstract}
 Fruit harvesting remains predominantly a labor-intensive process, motivating the development of research for robotic grippers. Conventional rigid or vacuum-driven grippers require complex mechanical design or high energy consumption. Current enveloping-based fruit harvesting grippers lack adaptability to fruits of different sizes. This paper introduces a drawstring-inspired, cable-driven soft gripper for versatile and gentle fruit harvesting. The design employs 3D-printed Thermoplastic Polyurethane (TPU) pockets with integrated steel wires that constrict around the fruit when actuated, distributing pressure uniformly to minimize bruising and allow versatility to fruits of varying sizes. The lightweight structure, which requires few components, reduces mechanical complexity and cost compared to other grippers. Actuation is achieved through servo-driven cable control, while motor feedback provides autonomous grip adjustment with tunable grip strength. Experimental validation shows that, for tomatoes within the gripper’s effective size range, harvesting was achieved with a 0\% immediate damage rate and a bruising rate of less than 9\% after five days, reinforcing the gripper’s suitability for fruit harvesting.
\end{abstract}

\section{Introduction}
\label{intro}

\begin{figure}[!t]
    \centering\includegraphics[width=0.85\columnwidth]{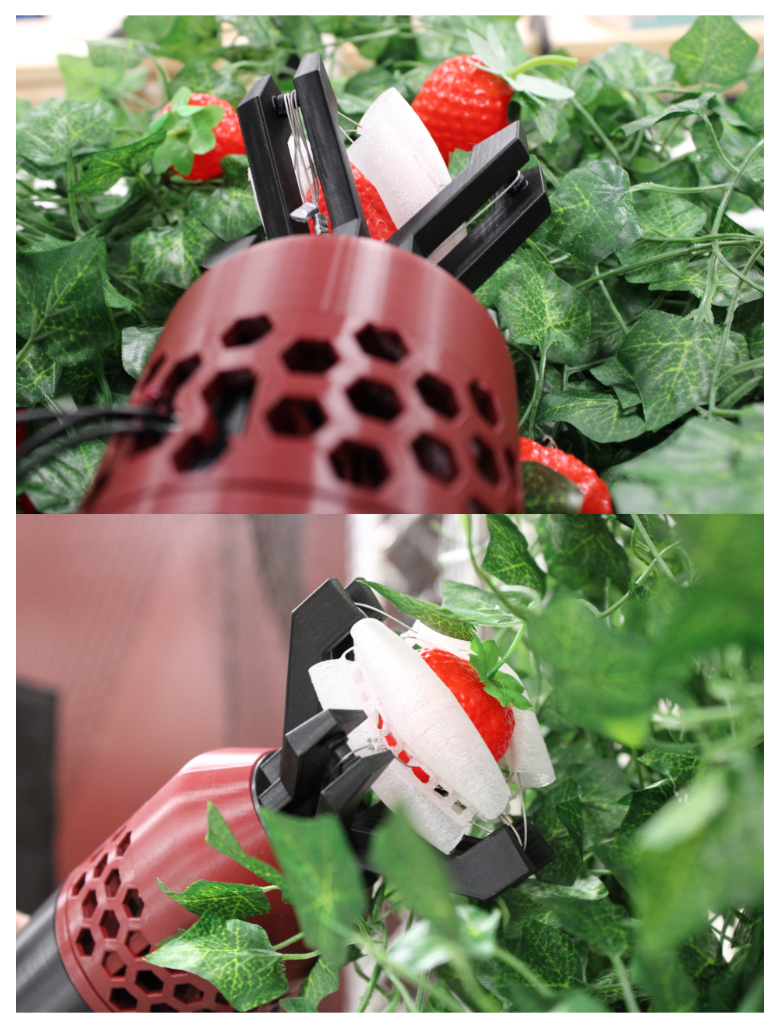}
    \caption{\textbf{Visualization of the ANGEL Gripper on Artificial Strawberries.} The main body of the gripper is fabricated with 3D-printed PLA. The contact surfaces are 3D printed with Thermoplastic Polyurethane (TPU). The gripper imitates the mechanism of a drawstring bag, using two motors to pull the cables routed through the pockets to control its opening and closing. This visualization explains how it works on strawberry models.}
    \label{gripper}
\end{figure}

While there is ongoing research and development towards fruit harvesting solutions\cite{zhang2020technology}\cite{vougioukas2025mechanization}, hand-picking remains the dominant method due to its delicacy for soft fruits \cite{calvin2010us}. The fruit production industry is primarily reliant on manual labor. 
However, rising costs and a shortage of labor pose issues with continued hand-picking in the industry. Certain fruits like strawberries tend to be more labor-intensive crops to cultivate \cite{calvin2022supplement}. Harvest costs account for up to 67 percent of total costs in the case of strawberry production in California \cite{tourteStrawberry}.
As a result, robotic harvesting methods have been a continuing target of research, aiming to automate fruit harvesting while limiting damage imparted onto the fruit. A particular area of concern is a robot's ability to adapt to natural variation in fruit size and shape \cite{wang2024review}. Without sufficient adaptability, the fruit is more likely to sustain bruising or damage, devaluing it \cite{kootstra2021selective}.      

The possibility of severe damage like bruising is a prevalent issue reported during fruit harvesting\cite{bruiseDamage}. These damages reduce the cosmetic appeal of fruits, discouraging consumers from purchasing them. Additionally, bruising can lead to mold growth, which can contaminate surrounding fruits during post-harvest \cite{moldPostHarvest}. To overcome this problem, researchers have turned to robotic grippers as a viable solution. This is because of their controllability and optimization towards a task, making them highly effective in harvesting environments.

However, many robotic grippers have rigid designs that fail to adapt to the fruit's natural variety in size and shape. They may also cause damage to fruits due to concentrated pressures on small areas of fruit surfaces \cite{uppalapati2020berry,EricACC,EricAIM}. To tackle this, robotic grippers ideal for fruit harvesting should have gentle, adaptive harvesting capabilities, making soft robotic grippers a suitable choice \cite{kootstra2021selective}. Nevertheless, contemporary soft robotic grippers still come with limitations. For instance, soft-fingered grippers that use sensors \cite{tendon-drivenGripper} require more maintenance, increase design complexity, and add fabrication costs in comparison with sensor-less soft grippers. 
Alternative designs, such as fingerless soft robotic grippers that envelop fruits, have also been explored. Elfferich et al. \cite{BerryTwist} introduced a gripper that envelopes a blackberry by twisting a tube of fabric around the fruit. However, their assembly requires many components like springs, rings, and gears, thereby adding mechanical complexity.
Other soft, enveloping gripper designs require multi-step silicone casting \cite{vaccumOrigami}. Such grippers are also vacuum-driven, which requires high power consumption.

\begin{figure*}[htbp]
\centering
\includegraphics[width=0.85\textwidth,trim=0 100 0 50, clip]{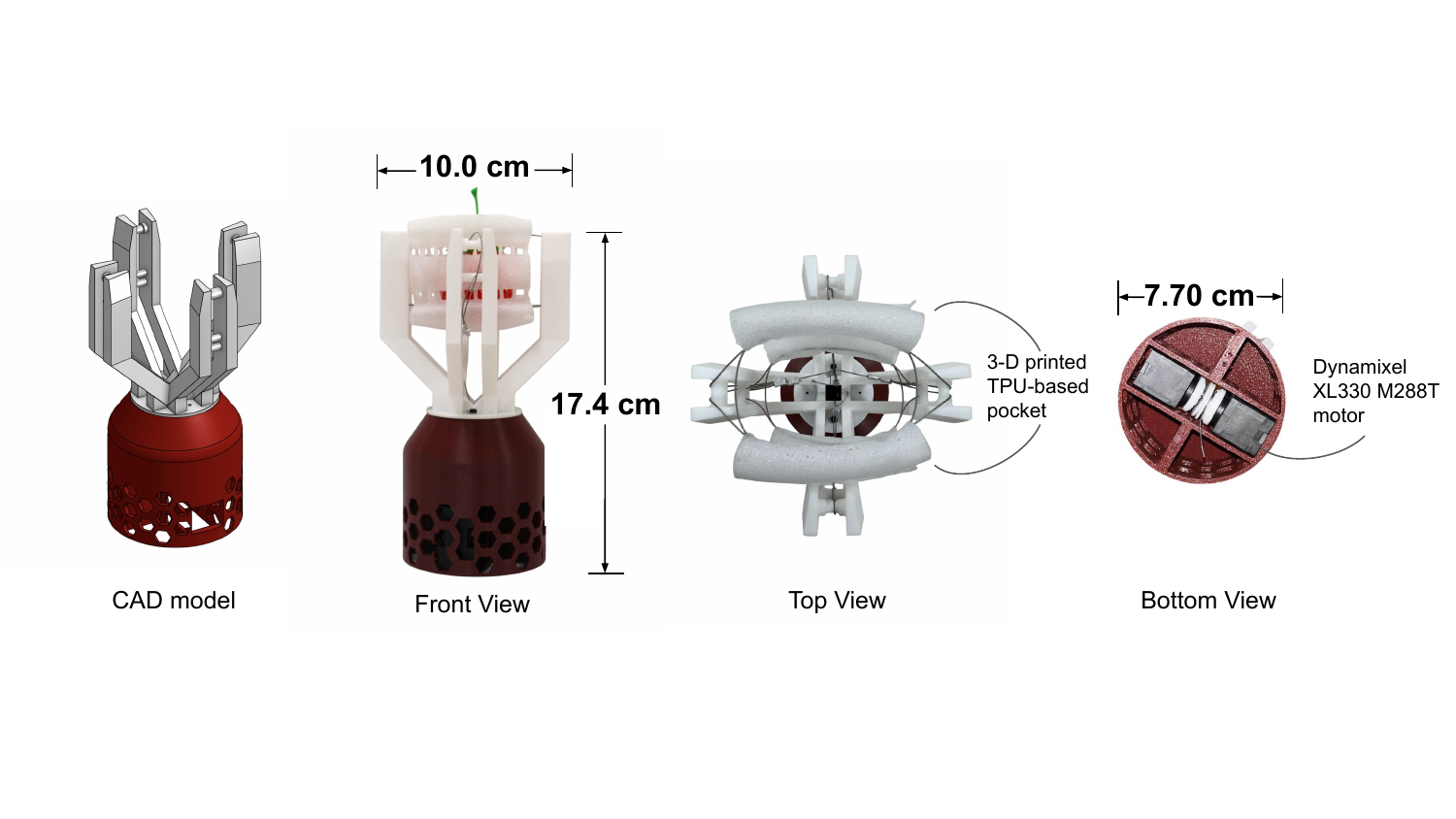}
\caption{\textbf{Gripper Components and Dimensions.} Visualized from left to right: the CAD model of the main body of the gripper, the front view with TPU pockets, the top view when the pocket envelope is open, and the bottom view with the Dynamixel motors installed.}
\label{fig:gripper_render}
\end{figure*}

In this paper, we propose a soft, cable-driven gripper suitable for gentle fruit grasping. The gripper imitates the mechanism of a drawstring bag, using a pair of Thermoplastic Polyurethane (TPU) pockets with steel wires routed through them, as depicted in Fig. \ref{fig:gripper_render}. As the wires are pulled, they close the gap between the pockets by forcing them inward, enveloping an object inside. The flexible material of the pockets conforms to the fruit's surface, applying a distributed force across the fruit. To open the gripper, a pair of wires pulls each pocket outward to accept an object, once the closing mechanism is released. 

Our contribution can be summarized as follows:
\begin{itemize}
    \item The paper introduces a drawstring-inspired, cable-driven soft gripper made from 3D-printed TPU pockets with integrated steel wires, achieving gentle and adaptive grasping with low mechanical complexity and reduced energy consumption compared to existing soft grippers.
    \item Experimental validation shows that for fruits within the gripper’s effective size range, it achieves a high harvest success rate with a 0\% immediate damage rate and less than 9\% minor bruising rate after five days.
\end{itemize}

\section{Related Works}
\label{related_works}

\begin{table*}[ht]
\centering
\caption{\textbf{Comparison of Grippers for Fruit Harvesting.}}
\label{tab:gripper_comparison}
\renewcommand{\arraystretch}{1.3}
\begin{tabularx}{0.8\textwidth}{|X|X|X|}
\hline
\textbf{Paper / Gripper} & \textbf{Mechanical Complexity} & \textbf{Control Requirement} \\
\hline
Johnson et al. \cite{johnson2024field} & Low ($\sim$8 components, structured 3D printed model) & Low (open-loop pneumatic inflation/deflation) \\
\hline
Zhao et al. \cite{zhao2025self} & High ($\sim$15 components, sensor integration, customized finger and body design) & High (electropneumatic control, tactile sensors, closed-loop adaptive feedback) \\
\hline
Gunderman et al. \cite{tendon-drivenGripper} & High ($\sim$16 components, sensor integration, customized finger and body design) & Medium-High (tendon actuation + fingertip force feedback) \\
\hline
BerryTwist \cite{BerryTwist} & Medium ($\sim$15 components, structured 3D printed models) & Low (open-loop with position control) \\
\hline
Gao et al. \cite{gao2022development} & Medium ($\sim$9 components: clamping finger, rotating and telescopic cylinders, gears, housing) & Medium (open-loop pneumatic control with RGB-D vision) \\
\hline
Ranasinghe et al. \cite{softPneumatic} & Medium ($\sim$10 components: doughnut-shaped actuators, pneumatic lines, pressure relief valves, pressure regulators) & Medium (open-loop pneumatic control for force regulation) \\
\hline
Ansari et al. \cite{ansari2024novelapproachtomatoharvesting} & High ($\sim$14+ components: six auxetic fingers, rigid exoskeleton, separator leaves, latex basket, Scotch-yoke drive, cutter with servo, multiple 3D-printed parts) & High (servo motor actuation with torque–force modeling, micro-servo pedicel cutter, RGB-D input) \\
\hline
\textbf{ANGEL Gripper (ours)} & Low (9 components, structured 3D printed model) & Low (current control, cable driven) \\
\hline
\end{tabularx}
\end{table*}

Existing harvesting mechanisms can be generally categorized into suction-based grippers, fingered mechanical grippers, enveloping-based grippers, and a hybrid of those designs.

Suction-based grippers are widely used in apple harvesters due to their ability to detach fruit with minimal mechanical contact. Zhang \cite{zhangJFR} and Hua \cite{HuaVacuumApple} both similarly integrated vacuum-cup end-effectors into a harvesting system and demonstrated their harvest success in orchards. 
There also exist grippers of hybrid designs leveraging soft materials for their adaptation to complex objects, suction capability, and unique attachment/detachment systems \cite{hao2020multimodal, SoftGripperforImprovedGraspingwithSuctionCups, vaccumOrigami, cucumberGripper, wang2023development}.
For example, Li \cite{vaccumOrigami} designed a vacuum-driven soft gripper made of an origami “magic-ball” and a flexible thin membrane. It demonstrated the ability to lift objects of various weights and types; however, no field test was conducted to evaluate its effect on fruits.
As for applications in fruit harvesting, Velasquez \cite{velasquez2024compact} designed a gripper that utilized compliant suction cups to gently attach to the fruit before deploying fingers to secure a grip. The gripper demonstrated its practical effectiveness in an apple orchard.
While effective for fruits like apples with a relatively hard surface, suction methods are energy-intensive and less suited to smaller, softer crops.

To address these limitations, specialized grippers have been developed to handle delicate fruits. 
In an early work by Dimeas \cite{dimeas2015design}, a finger-like gripper equipped with sensor arrays on each finger was designed for strawberry harvesting. A hierarchical control scheme was specifically proposed based on a fuzzy controller for force regulation.
It is also worth noting that Xiong et al. in \cite{xiong2020autonomous, XIONG2019392} developed a complete robotic system equipped with cable-driven grippers for strawberry harvesting. A notable aspect is that the internal container inside the gripper collects berries during picking, eliminating the need for the manipulator to travel back and forth to a separate punnet.
Recent efforts have transitioned to applying soft robotic grippers in handling delicate fruits.
Gunderman \cite{tendon-drivenGripper}, for instance, proposed a tendon-driven soft robotic gripper for blackberry harvesting. The gripper was composed of three soft silicone fingers. Each finger tip was equipped with a force sensor for proper force application upon grasping fruits. 
Similarly, Zhao\cite{zhao2025self} designed a soft pneumatic gripper system with embedded tactile sensors and a motorized syringe actuation mechanism that autonomously adapted to fruit size, grip force, and pressure. 
Elfferich et al. \cite{BerryTwist} introduced the BerryTwist, a gripper that enveloped a blackberry by twisting a tube of fabric around the fruit. However, their assembly required many components like springs, rings, gears, and cloth. A similar cylindrical pneumatic gripper in \cite{johnson2024field, softPneumatic}, was designed and validated with field experiments to harvest blackberries and strawberries.

Though the integration of extra force sensors and mechanical components could help alleviate damage to fruits, our proposed design seeks to reduce mechanical complexity, cost, and control requirements while minimizing fruit damage as compared in \cref{tab:gripper_comparison}.

\section{Materials and Designs}
\label{methods}

\begin{figure*}[ht]
\centering
\includegraphics[width=0.75\textwidth,trim=0 50 0 50, clip]{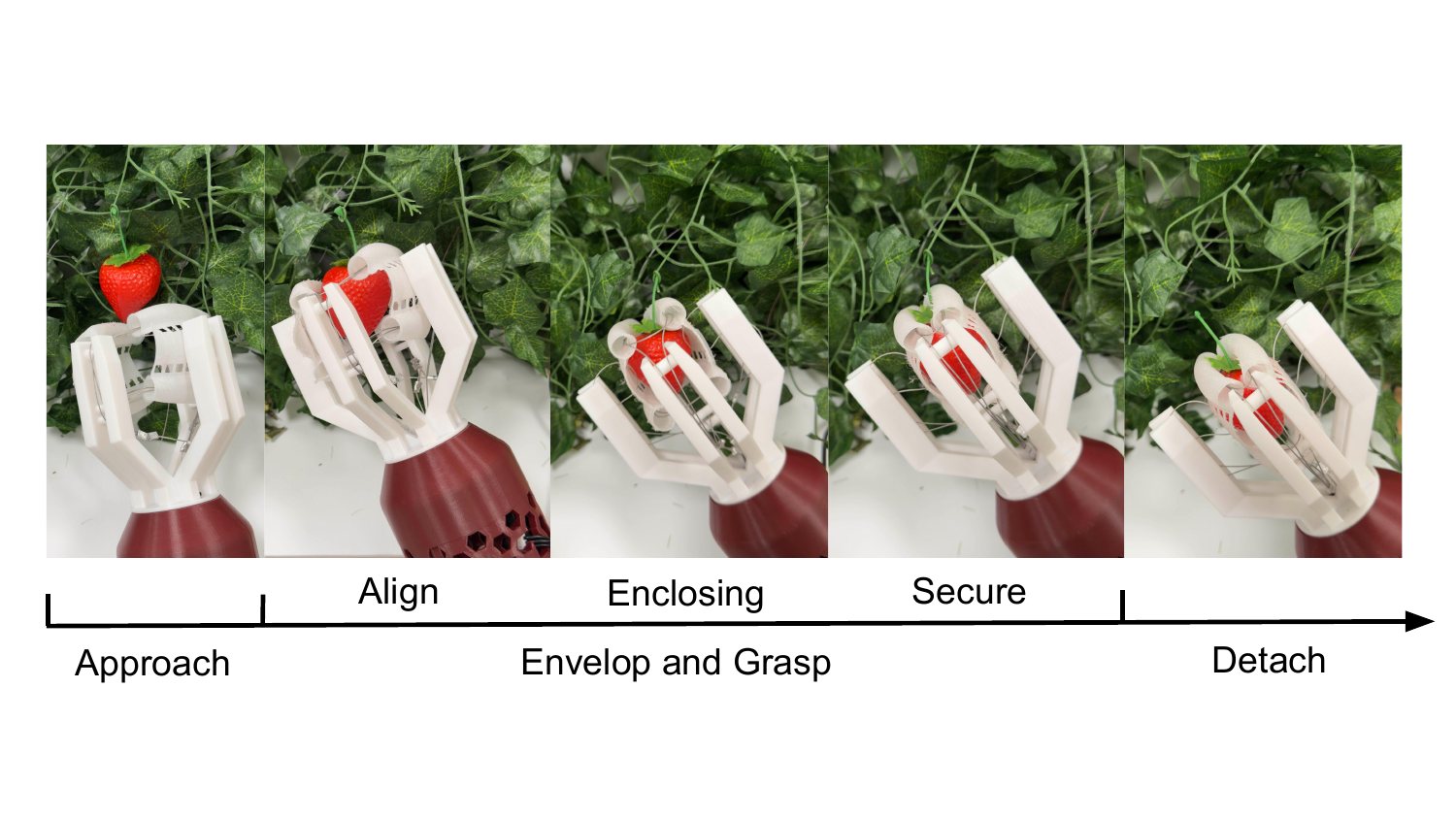}
\caption{\textbf{Harvesting Process Explanation and Visualization on Artificial Strawberries.}  From left to right, the gripper sequentially performs the steps of approaching, aligning, enclosing, securing, and detaching to complete the harvest. 
}
\label{fig:3_harvesting_process}
\end{figure*}

The goal of our gripper design is to enable a secure grasp of a fruit, adaptable to different sizes and shapes. We also aim to minimize damage sustained by the fruit. The gripper's design should be lightweight, require few parts, and have low control requirements.

\subsection{Materials and Fabrication}

\begin{figure*}[ht]
    \centering\includegraphics[width=0.8\textwidth]{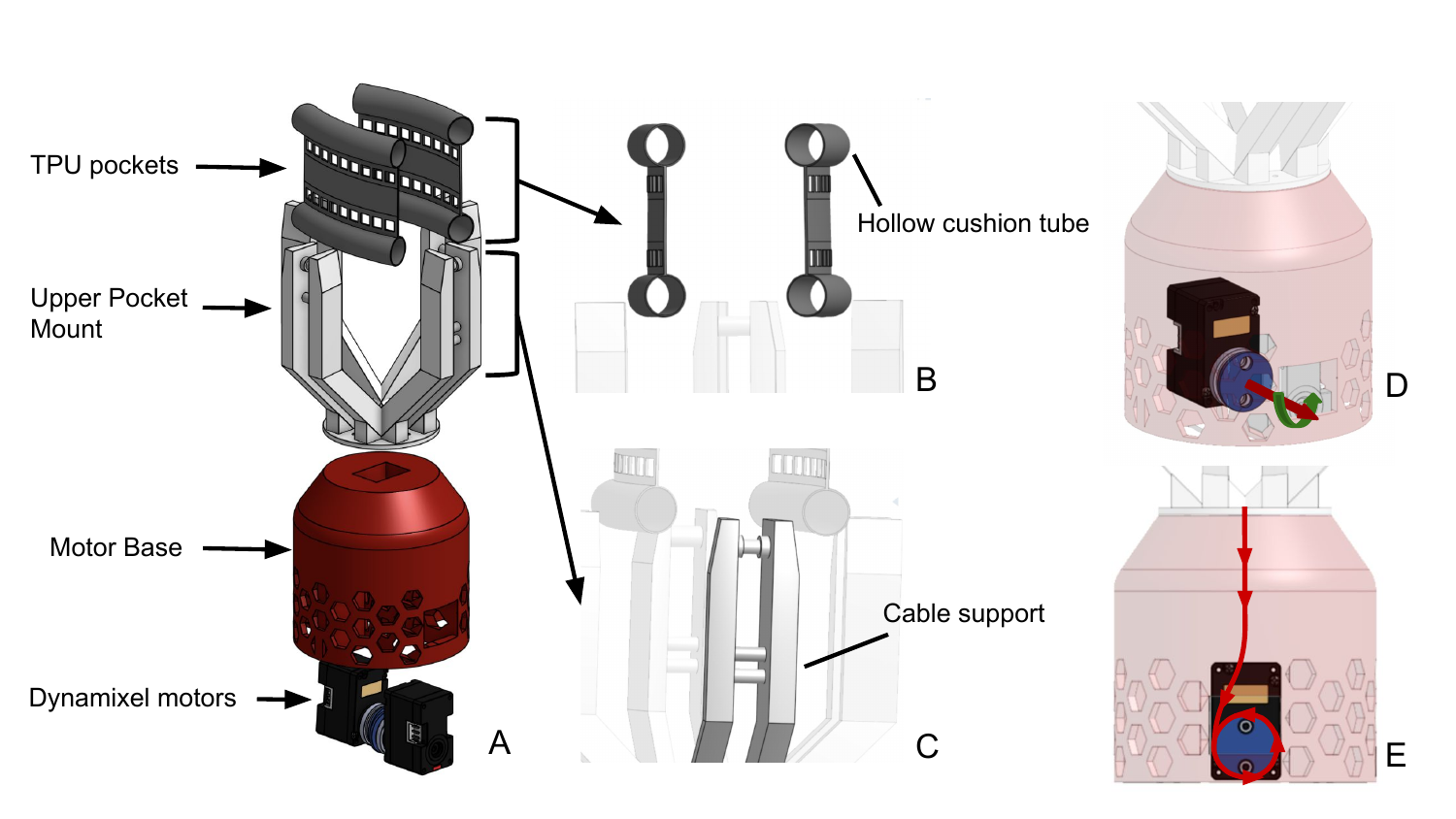}
    \caption{\textbf{Detailed View of the Gripper.} Subfigure A shows an exploded view of the gripper. Subfigure B illustrates the hollow-tube design of the pockets, which allows wiring and provides a cushioning layer between the pockets and the fruit. Subfigure C presents the cable support structure, which guides cable motion and ensures that the pockets are pulled in the intended direction. Subfigures D and E detail the winding of the steel cable around the motor spool, enabling both pulling and release. Note that cables are not visualized in the figure.} 
    \label{fig: exploded view}
\end{figure*}

Our design comprises 9 components: two fabricated TPU pockets, an upper pocket mount, steel cable, a motor base, two Dynamixel XL330-M288T motors, a circuit board mount, and a U2D2 Power Hub board. Refer to Fig. \ref{fig: exploded view} for a detailed view of the gripper design.

The gripping surfaces consist of a pair of 3D-printed TPU-based pockets. TPU 95A (shore hardness of 95 A) was used to fabricate the pockets, chosen for its flexibility. Each pocket includes two curved cushion tubes with perforated and solid sections binding the tubes. The perforations reduce the stiffness of the pockets, allowing better flexure.
The gripping pockets are designed with a curvature of 35 degrees, which enables a greater contact area and more uniform distribution of their applied forces. Each tube has a 6.5 mm outer diameter and an arc length of 6.1 cm. The top and bottom of the TPU pockets are fitted with the cushion tubes to reduce fruit damage and prevent the fruits from slipping out during harvesting. The pocket wall thickness is 0.4 mm, resulting in a lightweight structure with a total mass of only 3.94 grams.
\begin{figure}[t]
    \centering\includegraphics[width=0.85\columnwidth]{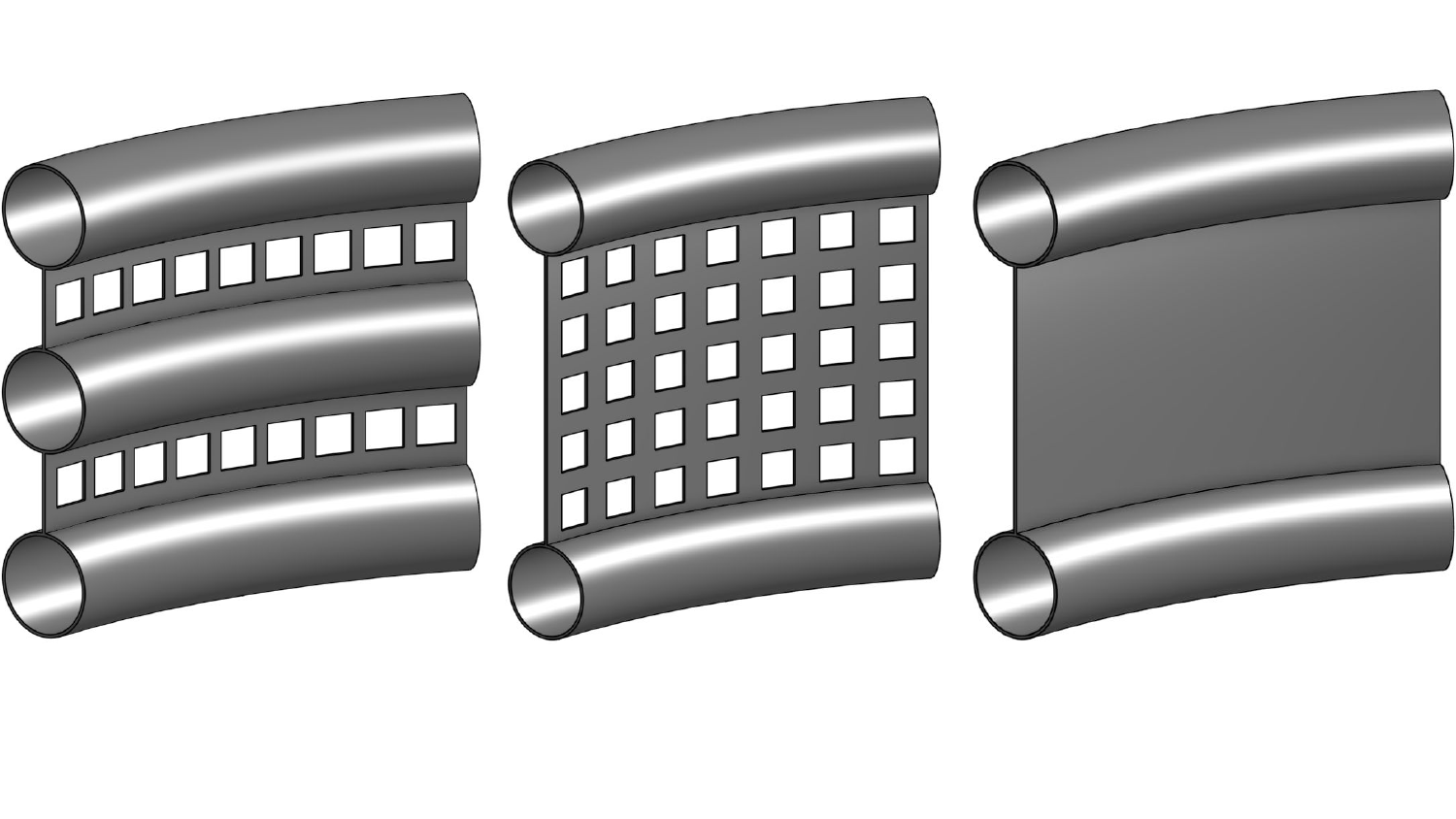}
    \caption{\textbf{Alternative Pocket Designs.} From left to right: TPU pockets optimized for cushioning, flexure, and contact area.} 
    \label{fig:pocket designs}
\end{figure}

As the pockets are easily printable with TPU, they may be modified to optimize the grasp of certain objects. To allow a softer touch, the number or diameter of the cushioning tubes can be increased. For higher contact area, a thin, solid fill can be applied between the tubes. Higher flexure can be achieved with perforations along the entire contact area. Fig. \ref{fig:pocket designs} illustrates examples of these pocket geometries.

The steel wires are routed through the pockets, and all of the ends are joined together using aluminum crimps, such that pulling one cable can operate the closing mechanism. 

The same is done with the opening mechanism, resulting in two cables that can open or close the pockets. These cables are each wound around a servo motor spool, allowing them to be actuated. Two Dynamixel XL330 M288T motors are used for the actuation of our gripper. This model of servo motor is capable of accurate position, velocity, and current control.  

The upper pocket mount is made of PLA (Polylactic Acid) material. The purpose of the mount is to guide the cables from the motors through the inside of the TPU pockets. The cables are supported with beams on each prong for better routing and grip geometry, illustrated in Fig. \ref{fig: mechanism}.
Additionally, the motor base is also made from PLA material. The purpose of the motor base is to hold the motors in place while a force is being applied upwards from the tensioned wires.

A PLA-printed mount houses the gripper’s electronic components. The Dynamixel motors are interfaced through a U2D2 Power Hub and control board, which provides communication over the Dynamixel protocol. Control commands (i.e., current) are issued from Python programs, such as our autonomous grasp routine, while real-time motor states (current, velocity, etc.) are monitored and plotted.

\subsection{Mechanism}
\label{subsec: mechanism}

\begin{figure}[t!]
    \centering
    \begin{subfigure}[t]{0.48\columnwidth}
        \centering
        \includegraphics[width=\columnwidth]{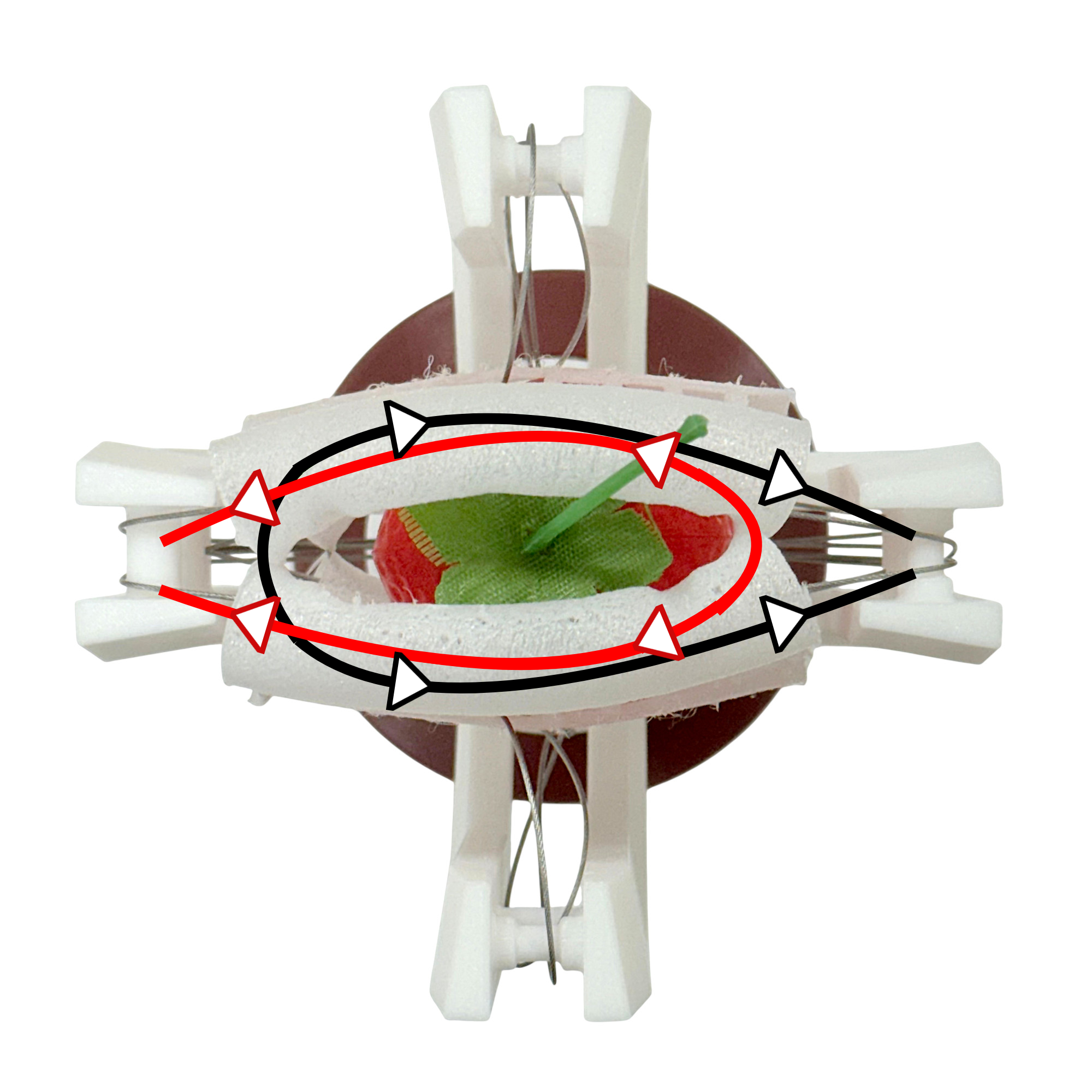}
        \caption{Closing Mechanism}
        \label{fig:subfigure a}
    \end{subfigure}
    \hfill
    \begin{subfigure}[t]{0.48\columnwidth}
        \centering
       \includegraphics[width=\columnwidth]{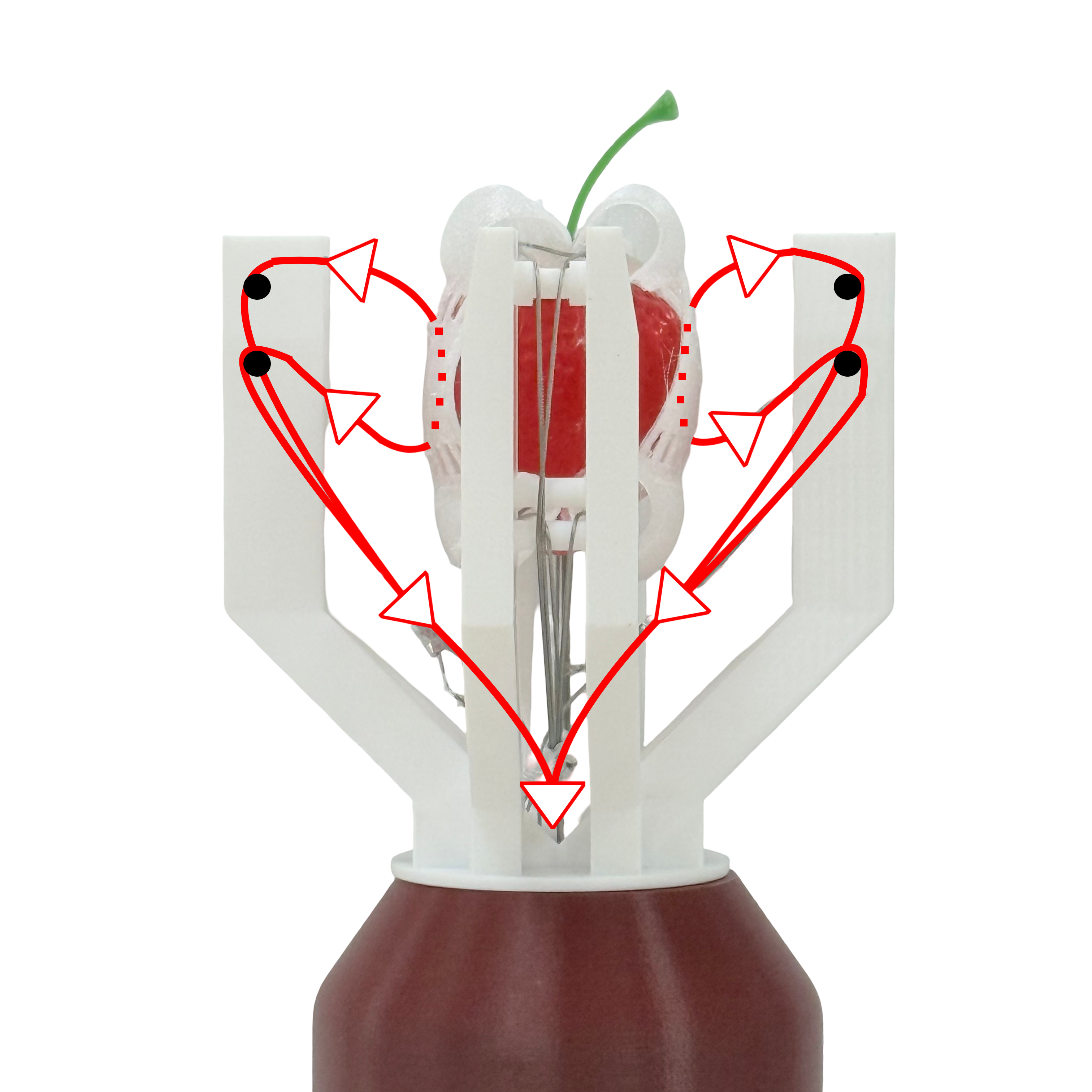}
        \caption{Opening Mechanism}
        \label{fig:subfigure b}
    \end{subfigure}
\caption{\textbf{(a) Closing Mechanism:} The cables, marked in black and red, are pulled by a motor in the direction of the arrows, bringing the pockets together to close. \textbf{ (b) Opening Mechanism:} The black dots denote cable supports that guide the directions of the cables. To open the pockets, a motor pulls a cable downward, pulling both pockets apart.}
\label{fig: mechanism}
\end{figure}

In this subsection, we explain the mechanism to open and close the pockets. Fig. \ref{fig: mechanism} visualizes the movement of the cables during the pocket closing and opening process. The whole harvesting process can be decomposed into five steps: approach, align, enclose, secure, and detach. This process is visualized in Fig. \ref{fig:3_harvesting_process}. 

\begin{figure}[t!]
    \centering
    \begin{subfigure}[t]{0.48\columnwidth}
        \centering
        \includegraphics[width=\columnwidth]{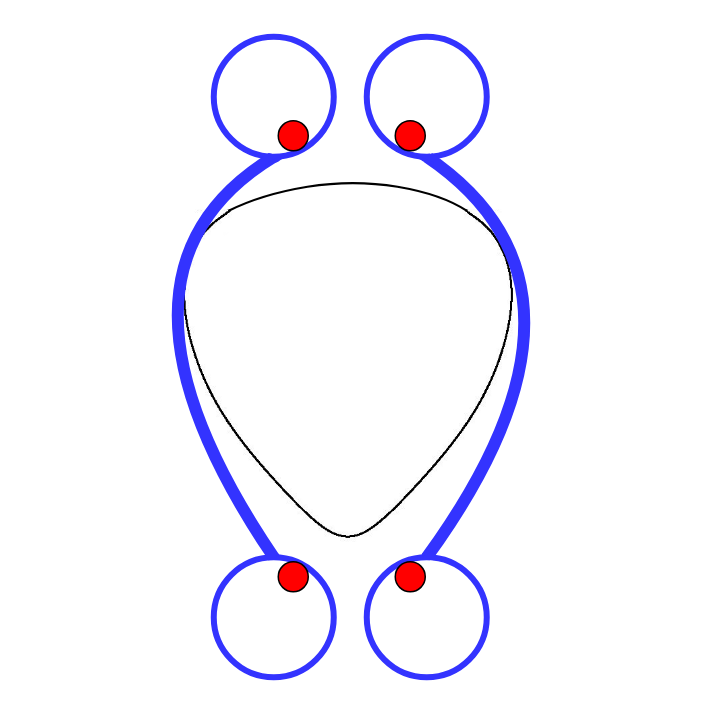}
        \caption{Grasp Geometry}
        \label{fig:subfigure a}
    \end{subfigure}
    \hfill
    \begin{subfigure}[t]{0.48\columnwidth}
        \centering
       \includegraphics[width=\columnwidth]{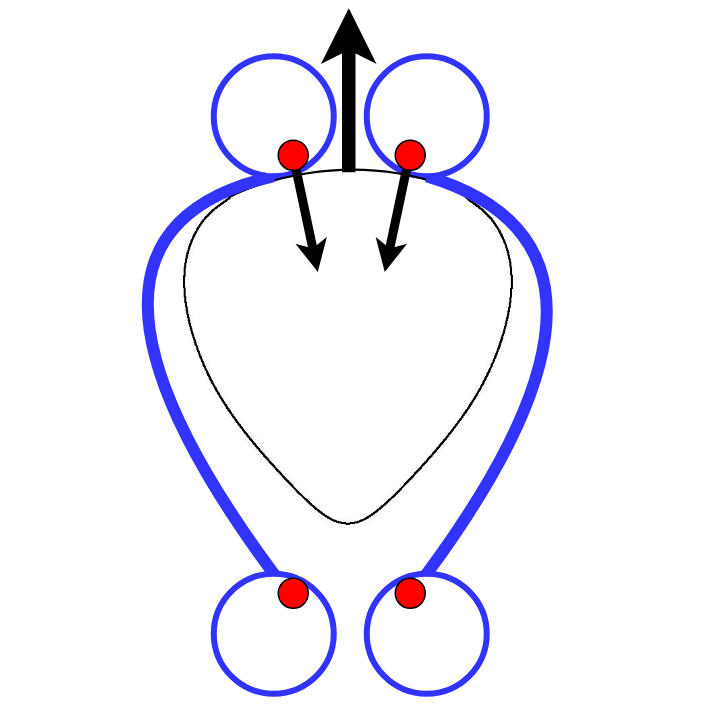}
        \caption{Free Body Diagram}
        \label{fig:subfigure b}
    \end{subfigure}
\caption{\textbf{Cross-sectional Diagram of a Fruit Grasp.} The conformity of the TPU pockets (blue) with the fruit surface is shown in \textbf{(a)}. Diagram \textbf{(b)} shows the opposing forces applied by the cables (red) when the fruit is pulled upwards.}
\label{fig: geometry}
\end{figure}

 The gripper operates on the same principle as a drawstring bag: when the cables are pulled, the circular opening tightens and the flexible pockets cinch inward around the object. As shown in Fig. \ref{fig: mechanism}, the linear pulling motion of the cables is redirected through the cushion tubes into a radial constriction of the pockets. Because the cable length inside the tubes is fixed, pulling them gathers the pocket edges toward the center. During closure, each cable pair naturally shifts toward the horizontal midpoint, constraining the object from the top and bottom. The upper cable segments prevent the object's upward movement; any attempt to pull the object out increases the downward force component of the upper cables, which actively resist its removal. The geometry of the grasp and corresponding forces are shown in Fig. \ref{fig: geometry}.
 
\begin{figure}[t!]
    \centering
    \begin{subfigure}[t]{0.48\columnwidth}
        \centering
        \includegraphics[width=\columnwidth]{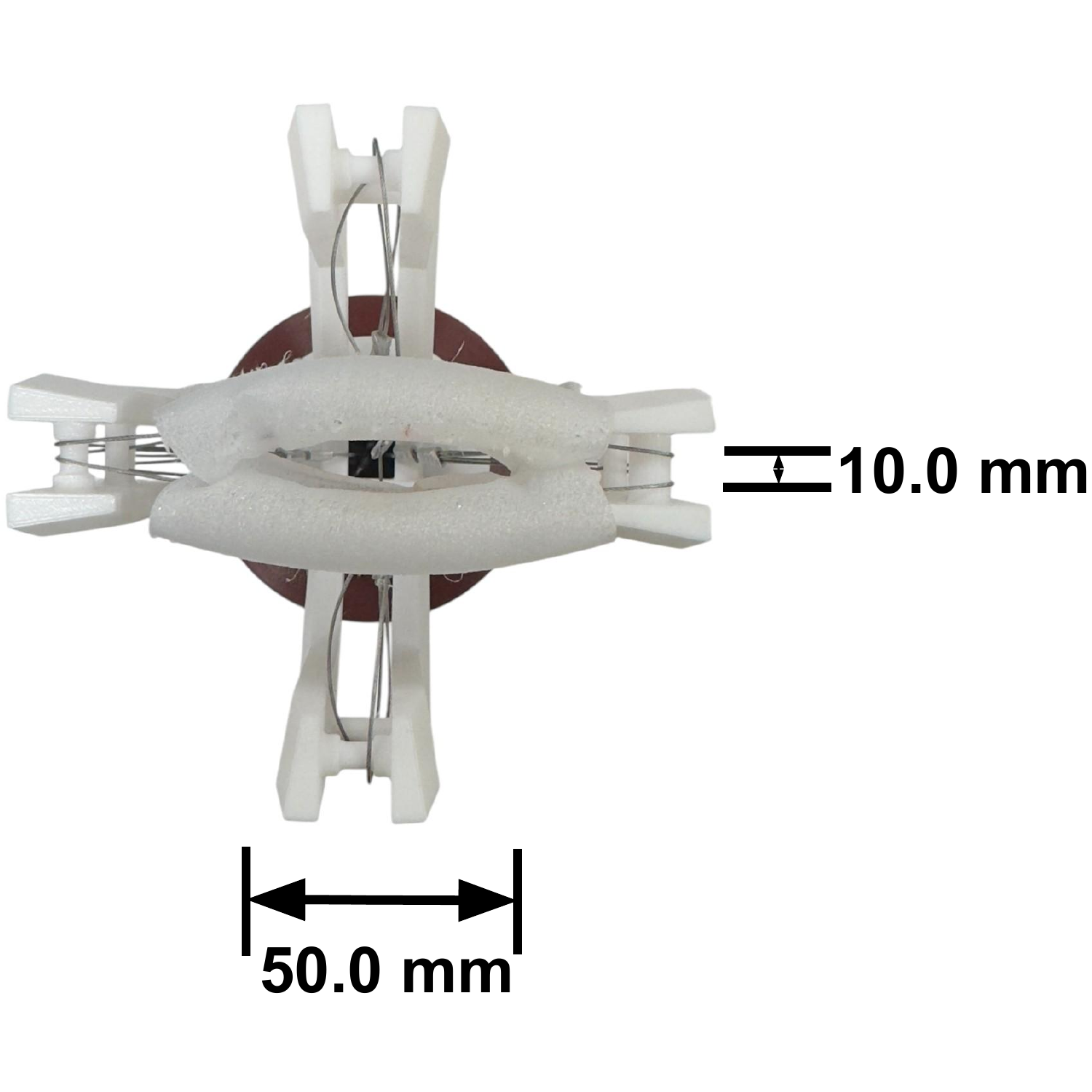}
        \caption{Minimal Fit-in Size}
        \label{fig:subfigure a}
    \end{subfigure}
    \hfill
    \begin{subfigure}[t]{0.48\columnwidth}
        \centering
       \includegraphics[width=\columnwidth]{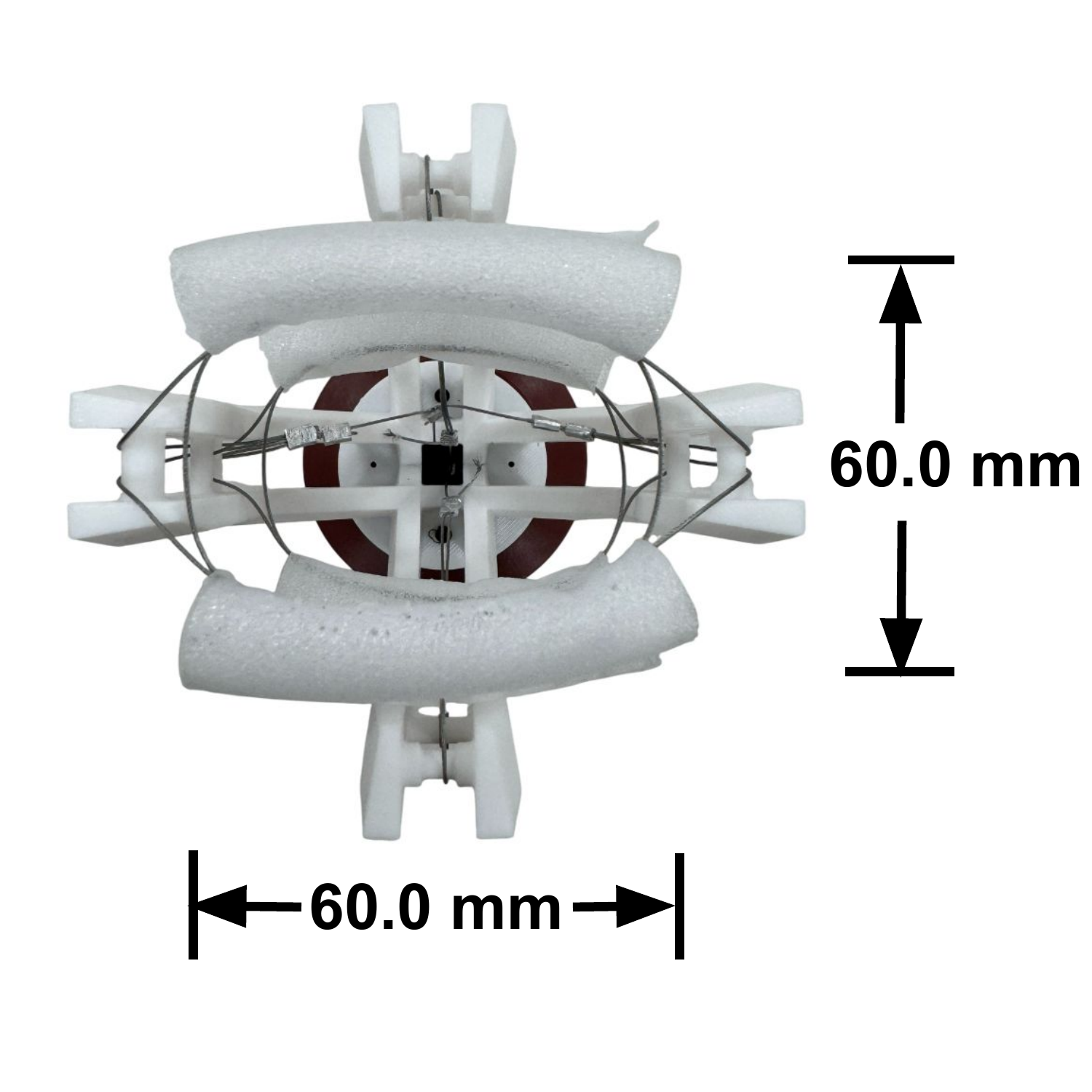}
        \caption{Maximal Fit-in Size}
        \label{fig:subfigure b}
    \end{subfigure}
\caption{\textbf{Fit-in Size of the Gripper}}
\label{fig: open_close_state_dimension}
\end{figure}

 In Fig. \ref{fig: open_close_state_dimension}, we measure the maximal and minimal fit-in size of fruits the gripper can grasp.

\subsection{Control}

\begin{figure}[t]
    \centering\includegraphics[width=\columnwidth]{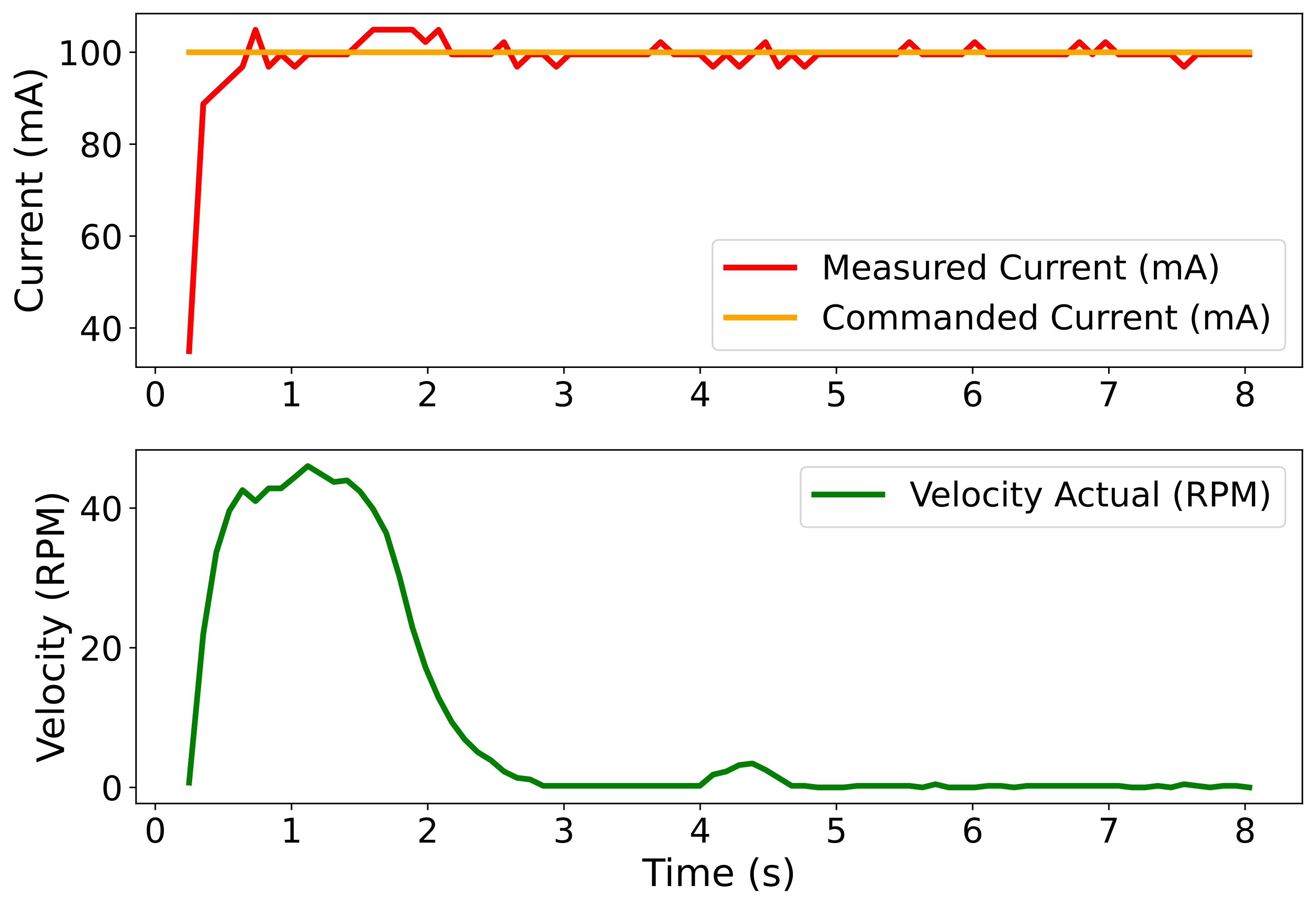}
    \caption{\textbf{Motor Current and Velocity Response During Grasping.}  The reference current is set to 100 mA. The current rises until reaching a steady-state value, and velocity decays to zero due to contact resistance.} 
    \label{fig: current_vs_time}
\end{figure}

The Dynamixel motors have a built-in current control loop, which adjusts their voltage duty cycle to achieve a target current draw. This is utilized to perform autonomous grasping by assigning a reference current for the motor to draw. The motor rotates when the current is applied, and slows down as resistance from the object in the pockets is encountered. At the steady state, the motor maintains the desired current without motion, indicating a grip has been attained. A higher reference current results in a tighter grip around the object. This behavior is demonstrated in Fig. \ref{fig: current_vs_time}: the upper plot shows the program reaching the 100 mA threshold, while the lower plot demonstrates the motor's rotational velocity gradually decreasing to zero as the object is gripped. 

A reference current of 100 mA was found to be optimal for most grasps. This value was determined by iteratively adjusting the reference current and evaluating the resulting grip strength and speed. An optimal grip was defined as one that could quickly and securely hold the object. Currents above 100 mA risked damaging fruits, while lower currents resulted in slower grasps.

\section{Experiments and Results}
\label{experiment}

\begin{figure*}[htbp]
\centering
\includegraphics[width=0.75\textwidth,trim=0 50 0 50, clip]{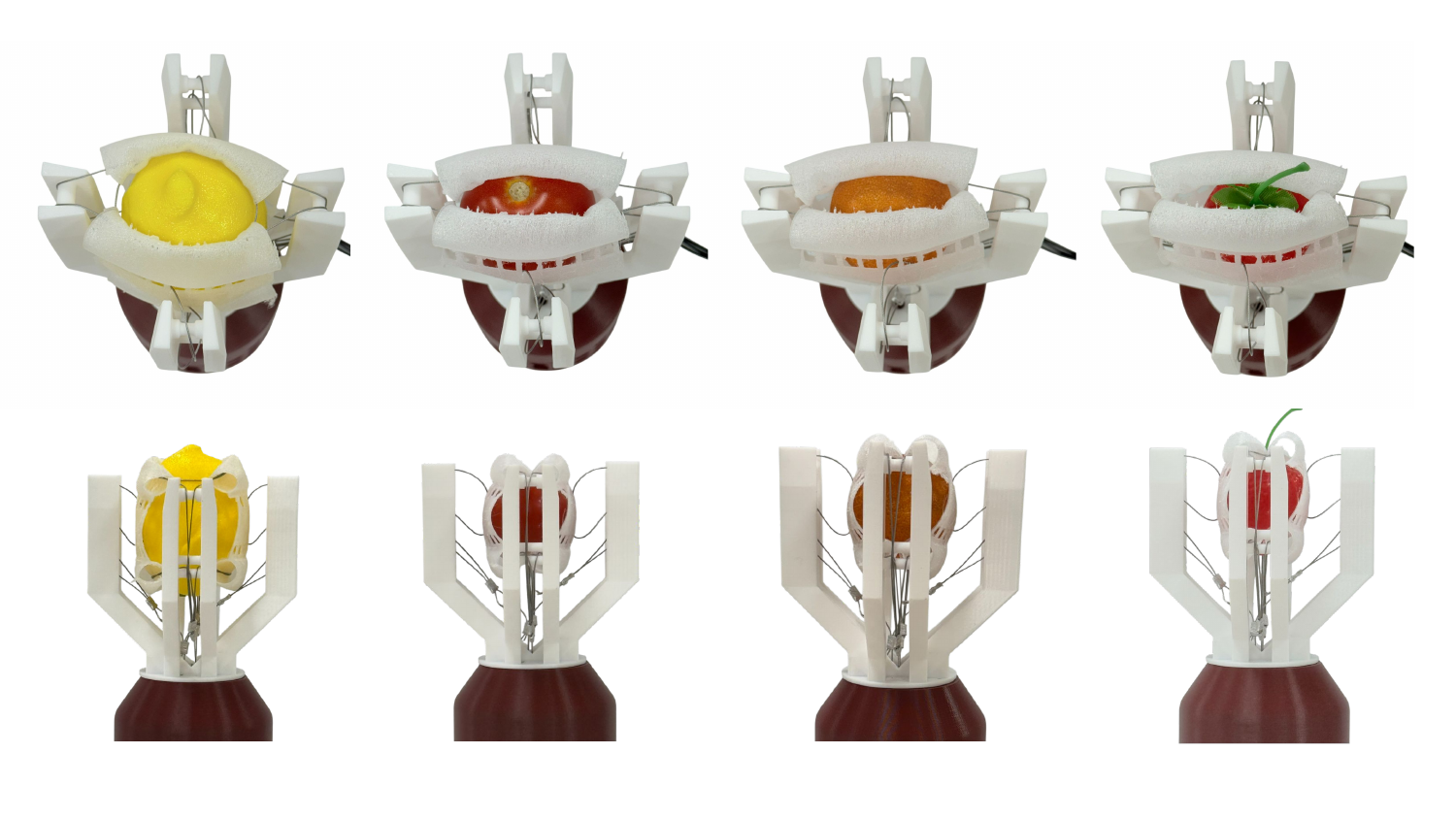}
\caption{\textbf{Gripper Operation on Different Fruits.} It shows the grasping of different kinds of fruits (from left to right: lemon, tomato, orange, strawberry) and demonstrates our gripper's adaptability to different fruit shapes and sizes.}
\label{fig:gripper_grasping_fruits}
\end{figure*}

In this section, we conducted a series of experiments aiming to answer the following questions:
\begin{enumerate}
    \item Is our gripper design capable of handling fruits of various shapes and sizes? \label{q:1}
    \item Can our gripper achieve a reasonable damage rate?
    \item Does our gripper reduce mechanical complexity and control requirements in comparison with grippers designed for similar purposes? 
\end{enumerate}
In each of the following subsections, we first describe the experimental design used to address the questions above, and then present and explain the results.
    
\subsection{Versatility}

To address question (1), we evaluate the proposed gripper from two perspectives: (a) its ability to grasp fruits of different sizes and shapes, and (b) its ability to pick up or harvest fruits successfully.

To validate the first aspect, the gripper was tested with fruits such as lemon, tomato, orange, and strawberry, with sizes ranging from 23 mm to 49 mm in diameter. As shown in Fig. \ref{fig:gripper_grasping_fruits}, the top and side views illustrate the gripper grasping and enveloping the fruits, providing a qualitative evaluation.

To demonstrate the second point, we simulated the harvesting process by pulling in-season fruits like tomatoes from vines.
Since the paper focuses on the design of a light-touch gripper rather than fruit harvest automation, the gripper was manually held to approach tomatoes and pull them from the vines, while the opening and closing of the pockets was automatically controlled by programming as described in \cref{subsec: mechanism}. We recorded the number of tomatoes harvested successfully without visible bruising during the process. 

We evaluated the gripper’s picking success rate using two groups of tomatoes of different sizes to ensure size variety during the test. The total number of tomatoes being tested was 56. The tomatoes were classified as medium- or small-sized based on their dimensions. The average diameter of the medium-sized tomatoes was 43.6 mm, while that of the small-sized tomatoes was 24.3 mm. All tomatoes used in the experiments were successfully picked without any visible damage or bruising upon harvesting. This high success rate can be attributed to the fact that the gripper’s performance largely depends on whether the fruits can fit into its pockets. In our experiments, the fruit sizes ranged from 21 mm to 51 mm, which falls within the gripper’s effective handling range.

\begin{table}[ht]
\centering
\caption{\textbf{Damage Rate}}
\label{tab: damage_rate}
\begin{tabular}{ c c c }
\hline
\rowcolor{gray!40}
                      & \makecell{Damage rate \\(upon harvesting)} & \makecell{Bruise rate \\(5 days after harvesting)}\\ \hline
\makecell{Medium sized tomatoes \\ ($n = 23$)}&           0\%                  &   8.6\%                                       \\ \hline
\makecell{Small sized tomatoes \\ ($n = 33$)} &           0\%                 &   9.0\%                                       \\ \hline
\end{tabular}
\end{table}

\subsection{Damage Rate}

To address question (2), we measured the damage rate of tomatoes during experiments in which they were pulled from their stems. We compared our results against reported damage rates in field harvesting. For machine-based harvesting, the damage rate varies among different methods, fruit maturity, and machine-specific designs, typically ranging from 20\% to 29\%, but can be reduced to 10\% with model-specific adjustments \cite{arazuri2010evaluation}. However, no damage rates and bruise rates have been reported for manual harvesting. 

\begin{figure}[!t]
    \centering\includegraphics[width=0.97\columnwidth]{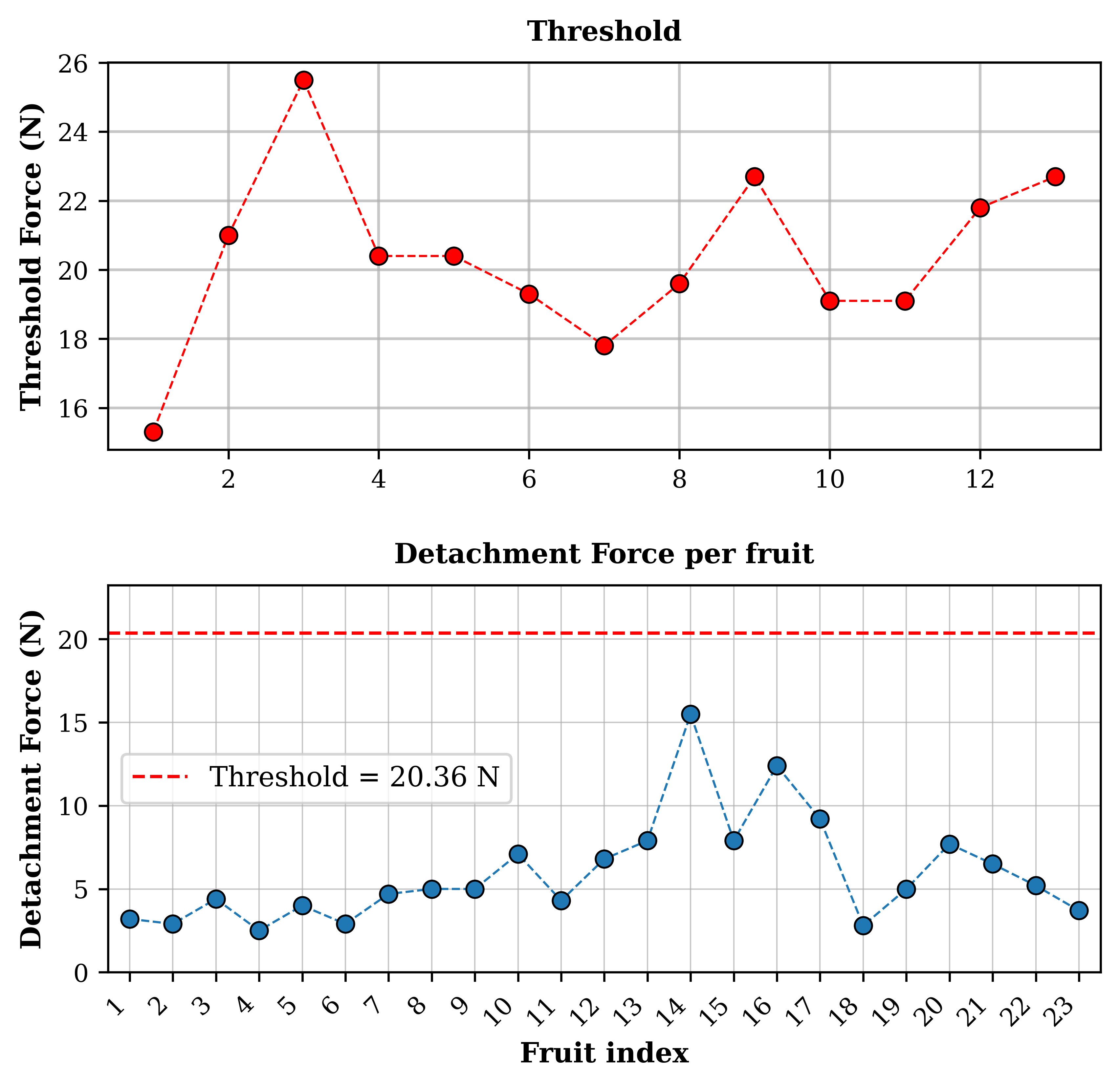}
    \caption{\textbf{Detachment Forces of Medium-Sized Tomatoes.} The top plot shows the threshold forces at which medium-sized sampled tomatoes begin to incur damage. The average threshold, calculated across 13 tomatoes, is used as the reference threshold. The bottom plot shows the detachment forces applied by the gripper, with the red line indicating the threshold.}
    \label{fig: detachment_force_medium}
\end{figure}

\begin{figure}[!t]
    \centering\includegraphics[width=\columnwidth]{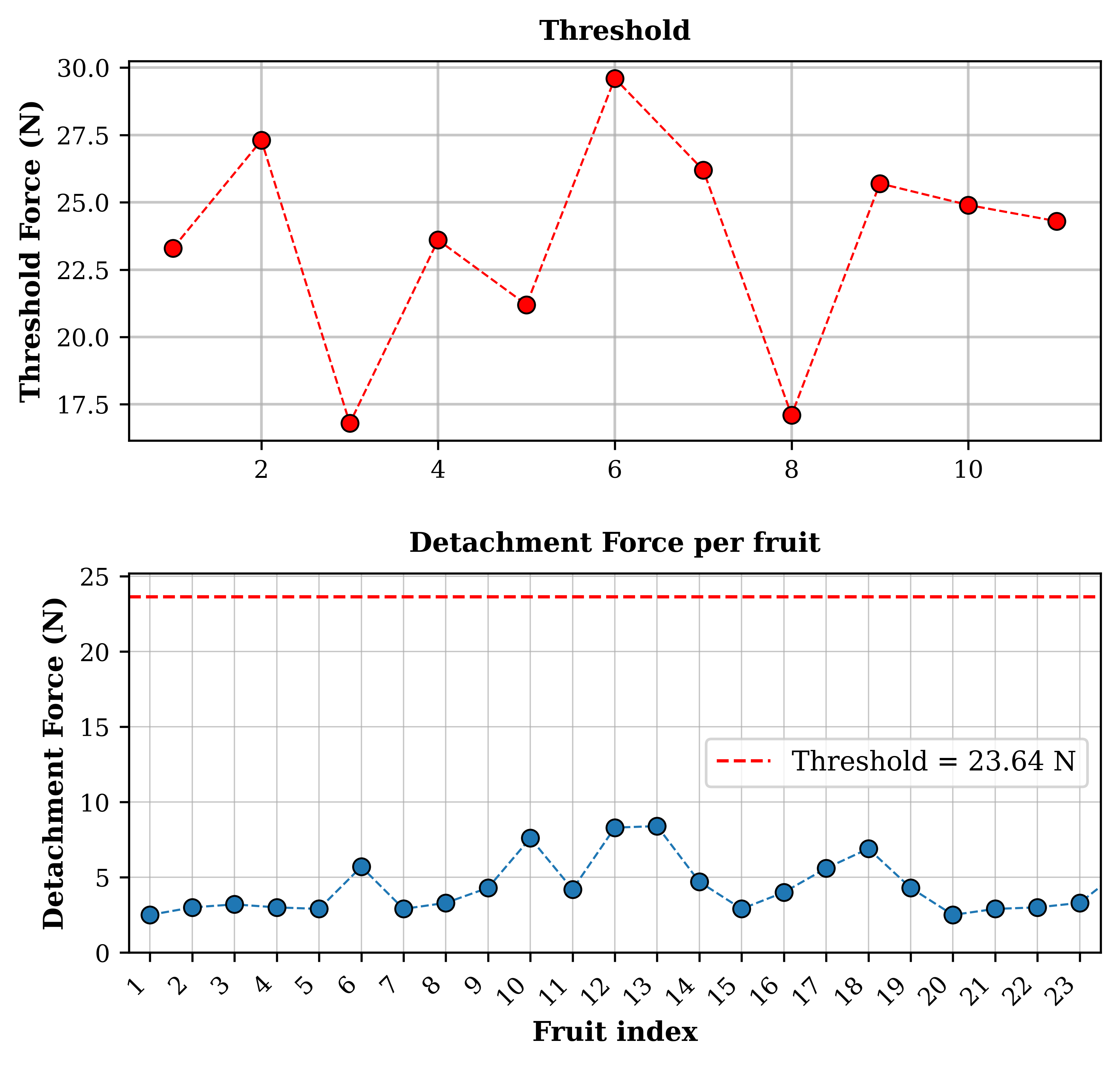}
    \caption{\textbf{Detachment Forces of Small-Sized Tomatoes.} The top plot shows the threshold forces at which small-sized sampled tomatoes begin to incur damage. The average threshold, calculated across 11 tomatoes, is used as the reference threshold. The bottom plot shows the detachment forces applied by the gripper, with the red line indicating the threshold.}
    \label{fig: detachment_force_small}
\end{figure}

We measure the damage rate in two ways. First, we compare the forces exerted on the fruits by the gripper with the threshold force required to cause immediate visible damage. To determine this threshold, tomatoes were intentionally pressed until bursts or irreversible bruises appeared, and the corresponding forces were recorded. A force sensor was used to measure the forces applied to the fruit surfaces from the top and side, simulating the contact conditions of the gripper. To detect the force exerted by the gripper on the fruit, a sensor was used to measure the peak vertical force required to pull the grasped tomato from its vine. In Fig. \ref{fig: detachment_force_medium} and Fig. \ref{fig: detachment_force_small}, we show the results of the measured threshold and applied forces by the gripper on the two groups of tomatoes. The top plots record the threshold forces needed to cause damage to the sampled tomatoes. We take the average of the measured forces as the threshold. As shown in the bottom plots of both figures, the forces exerted by our gripper remain well below the damage thresholds, with a substantial safety margin. These results provide quantitative evidence of the delicacy of the proposed gripper design.

Second, we inspect the fruits five days after harvest, since some bruises are gradually visible as time goes on.
Any fruit with indentation, flattening, or collapse on the surface is considered bruised.
Table \ref{tab: damage_rate} shows the damage rate upon harvesting and the bruise rate five days afterwards. The gripper achieves a damage rate of 0\% right after harvesting and a bruise rate of less than 9\% after 5 days. 
The bruised regions, measuring between 2 mm and 4 mm in length, did not lead to mold growth or fruit decay.
The damage rate observed is significantly lower than that typically reported for machine-based harvesting. Since there are no standardized benchmarks for bruise rates, we report these values here for the completeness of results.

\subsection{Mechanical Complexity and Cost}

Table \ref{tab:gripper_comparison} compares our design and other soft fruit grippers. The comparison considers mechanical complexity and control requirements. However, quantitative evaluation is challenging due to the lack of common benchmark designs or metrics, and the use of 3D-printed components in both their grippers and ours further complicates accurate cost estimation. Since detailed bills of materials are not reported in prior works, we classify each gripper into three relative levels (low, medium, high) for a high-level comparison. To evaluate mechanical complexity, we consider the number of components, the use of sensors, and the CAD design and assembly process as our metric. For control requirements, we compare their actuation methods and control schemes. In addition, we primarily compare against grippers designed for harvesting soft fruits to ensure a relatively fair comparison, since our gripper is also intended for handling delicate fruits.

As shown in Table I, our design requires fewer components and simpler control. Compared with the gripper proposed by Johnson et al. \cite{johnson2024field}, our design demonstrates similar mechanical complexity and control requirements, but excels in two aspects: (a) versatility, as it can handle fruits of varying shapes and sizes, as shown in Fig. \ref{fig:gripper_grasping_fruits} and (b) energy efficiency, as our design does not rely on pneumatic actuation but instead requires a maximum of 0.1 Watts of power during actuation.

\section{Conclusion}
\label{sec: conclusion}
This paper presents a drawstring-inspired gripper for fruit harvesting that combines TPU-based flexibility with a cable-driven mechanism to achieve gentle and adaptive grasping. The proposed design reduces both mechanical and control complexity compared to existing grippers. Experiments also demonstrate the effectiveness of the gripper in reducing the damage rate of tomatoes in comparison with machine-based harvesting.

The gripper also has some limitations. It is unable to grasp multiple fruits simultaneously, such as clusters of grapes or cherries. Further field experiments would be necessary to evaluate the gripper’s performance on softer fruits and a wider variety of crop types.
Future research will focus on fully automating the harvesting process and integrating the gripper into a complete robotic system. This would include the integration of computer vision, mobility, and attachment to a robotic arm.

\bibliographystyle{IEEEtran}
\bibliography{ref}

\end{document}